\documentclass{article}

\usepackage[final,nonatbib]{nips_2016}

\usepackage[utf8]{inputenc} 
\usepackage[T1]{fontenc}    
\usepackage{url}            
\usepackage{booktabs}       
\usepackage{amsfonts}       
\usepackage{nicefrac}       
\usepackage{microtype}      

\usepackage{color}
\usepackage{amsmath}
\usepackage{bm}
\usepackage{amsthm}

\usepackage{times}
\usepackage{graphicx} 
\usepackage{subfigure} 

\usepackage{algorithm}
\usepackage{algorithmic}

\usepackage{amssymb}
\usepackage{xr}
\usepackage{lscape}
\usepackage{dsfont}

\def\nn{\nonumber}

\def\E{\mathbb{E}}

\allowdisplaybreaks

\title{Unsupervised Learning of Predictors from Unpaired Input-Output Samples}

\author{
  Jianshu Chen, Po-Sen Huang, Xiaodong He, Jianfeng Gao and Li Deng\\
  Microsoft Research, Redmond, WA 98052, USA \\
  \texttt{\{jianshuc, pshuang, xiaohe, jfgao, deng\}@microsoft.com} \\
}

\begin{document}

\maketitle

\begin{abstract}
 
Unsupervised learning is the most challenging problem in machine learning and especially in deep learning. Among many scenarios, we study an unsupervised learning problem of high economic value ---  learning to predict without costly pairing of input data and corresponding labels. Part of the difficulty in this problem is a lack of solid evaluation measures. In this paper, we take a practical approach to grounding unsupervised learning by using the same success criterion as for supervised learning in prediction tasks but we do not require the presence of paired input-output training data. In particular, we propose an objective function that aims to make the predicted outputs fit well the structure of the output while preserving the correlation between the input and the predicted output. We experiment with a synthetic structural prediction problem and show that even with simple linear classifiers, the objective function is already highly non-convex. We further demonstrate the nature of this non-convex optimization problem as well as potential solutions. In particular, we show that with regularization via a generative model, learning with the proposed unsupervised objective function converges to an optimal solution. 
 
\end{abstract}

\section{Introduction}
Unsupervised learning, one major branch of machine learning involving learning without labeled data or without costly pairing input-output training data, has been a long standing research over decades. But it has achieved much less success compared with supervised learning that requires paired training data. Part of the difficulty in unsupervised learning is a lack of solid evaluation measures in the past. In this paper, we take a practical approach to grounding unsupervised learning using the same evaluation measure as that for supervised learning in prediction tasks without requiring paired input-output training samples. If successful, the benefit of such unsupervised learning would be tremendous. For example, in large scale commercial speech recognition systems, the currently dominant supervised learning methods typically require a few thousand hours of training material where each utterance in the acoustic form needs to be explicitly labeled with the corresponding word sequence by human. Although there are millions of hours of natural speech data available for training, labeling all of such acoustic data followed by supervised learning is simply not feasible. To make effective use of such huge amounts of acoustic data in speech recognition, the practical unsupervised learning approach outlined above would be called for.

In recent years, supervised learning has shown great successes in several major prediction tasks including speech recognition \cite{DNN_SPMagazine12, dahl2012context}, image recognition \cite{Krizhevsky_imagenet_NIPS12, zeiler2014visualizing}, machine translation \cite{Sutskever_Seq2Seq_NIPS14, Bahdanau_attention_ICLR15}, spoken language understanding \cite{mesnil2013investigation, mesnil2015using}, and image captioning \cite{Fang_caption_CVPR15, karpathy_caption_CVPR15, vinyals2015show, Xu2015show}. These successes rely heavily on training highly expressive deep learning models using large amounts of labeled training data. That is, the training examples are input-output pairs, where the outputs are labels obtained typically by costly manual annotations.  Unsupervised learning, however, is not as successful on these prediction tasks, although it has found other useful applications such as clustering \cite{xu2005survey}, text analysis \cite{Blei_LDA_JMLR03}, etc. The majority of the work on unsupervised learning for prediction tasks in the past has been to exploit the learned representations of the input data as feature vectors which are subsequently fed to a separate classifier; e.g., \cite{Le_Unsupervise_ICML12}. This approach, albeit widely used, is usually less effective than end-to-end learning with labeled data \cite{Chen_LDA_NIPS15}. Another important line of work on using unsupervised learning to help prediction is pre-training, where an unsupervised model trained using unlabeled data is used to initialize a separate supervised learning algorithm \cite{hinton2006reducing, Bengio_greedy_NIPS06, Bengio_deep_09, mikolov2013efficient, Dai_semisupervise_NIPS15}. Pre-training is shown to be effective only when there is a small amount of labeled data available \cite{DNN_SPMagazine12}. In prediction tasks with large amounts of paired training data, all the above unsupervised methods have played only an auxiliary role in helping supervised learning.

In this paper, we consider the unsupervised learning problem from a new and practical perspective. That is, instead of using unsupervised learning as an auxiliary step for supervised learning, we aim to develop an unsupervised learning algorithm that learns the input-to-output mapping (i.e., the predictor) from unpaired input-output training samples. Our approach has tremendous economic value in that it allows us to use a large amount of unlabeled data directly for prediction tasks. As we proceed to show in the paper, this is a very challenging problem since no clear and effective cost function has been established for such a problem in the literature. This paper represents our initial attempt to address this challenge by exploiting the \emph{sequence structure of the output samples} to learn the predictor. This is dramatically different from most previous work which often exploits the structure of the input samples. The objective function we defined aims to make the predicted outputs fit well the structure of the output (e.g., a sequence structure that is learned separately using only output samples), while preserving the correlation between the input data and the predicted output labels. We will give a detailed study of this objective function on a predictive task in order to understand the nature and difficulties of the problem, as well as its potential solutions. 

\section{Related Work}

For unsupervised learning applied to prediction and related tasks, several main approaches have been taken in the past. An important line of research has been to focus on exploiting the structure of input data by learning the data distribution using maximum likelihood rule. The most successful examples in this category include the restricted Boltzmann machine (RBM) \cite{Smolensky_RBM_1986, hinton2006reducing}, the deep belief network \cite{hinton_dbn_2006}, topic models \cite{Blei_LDA_JMLR03}, etc. The main technical challenge of these methods is the difficulty of computing the gradient of the likelihood function exactly. For this reason, various approximate methods have been developed, such as variational inference \cite{jordan1999introduction} and Monte Carlo methods \cite{hastings1970monte}.

Another important development is the methods that avoid the difficulties that arise in using maximum likelihood rule as the direct learning objective. These methods include autoencoder \cite{Bengio_deep_09}, denoising autoencoder \cite{vincent2010stacked}, variational autoencoder \cite{kingma2013auto}, and generative adversarial network (GAN) \cite{goodfellow2014generative}. However, these methods have been developed also aiming to model the input data distribution instead of learning the input-to-output mapping from unpaired input-output data.

A recent study that is more closely related to what we describe in this paper is \cite{sutskever2015towards}, which proposes the output distribution matching (ODM) as an alternative unsupervised learning objective to the likelihood function of the data. The ODM cost function measures how well the distribution of each predicted output sample matches the distribution of  target output samples. Dual autoencoder and GAN are used to implement the learning algorithm approximately. However, ODM does not exploit the structure of the output samples. In contrast, in the study reported in this paper, we explicitly exploit the sequence prior, a type of structure commonly found in speech and natural language data, of the output samples in the form of joint probability distribution of the outputs. We believe that the stronger the prior is, the better chance there is for this approach to work that exploits output distributions as the prior. The sequence prior is very strong, and in many possible applications such as speech recognition, machine translation, and image/video captioning, this sequence prior can be obtained from language models trained using a very large amount of text data freely available. The power of such a strong prior of language models in unsupervised learning has been demonstrated in an earlier study reported in \cite{Knight_ACL06}.

In addition to exploiting output distributions as the structured prior, our approach further exploits other sources of prior information including the correlation between input and output. The latter is implemented in our work as a regularization term of the objective function, which is derived from a generative model with information flow from output to input. The use of generative models in our work is similar to an earlier study reported in \cite{berg2013unsupervised} and to a more recent study reported in \cite{rasmus2015semi}.
Finally, our proposed unsupervised learning cost can be directly optimized using stochastic gradient descent in an end-to-end manner.

\section{Problem Formulation}

In this section, we first formulate the unsupervised learning problem. Let $x_t$ be $t$-th input vector, which is an $M$-dimensional real-valued vector, and let $y_t$ be the $t$-th output vector. In this paper, we consider the classification problem so that $y_t$ is a $C$-dimensional one-hot vector that represents one of the $C$ classes. In prediction tasks, the objective is to learn the conditional probability $p(y_t | x_t, W_d)$ from training samples, where $W_d$ represents the model parameter. $p(y_t | x_t, W_d)$ can be any parametric model such as neural networks.

In supervised learning problems, the training algorithm is presented with paired data $(x_t, y_t)$, which are assumed to be generated from a ground truth distribution $p(x_1,\ldots, x_T, y_1, \ldots, y_T)$. A common supervised training objective is
	\begin{align}
		\max_{W_d} \sum_{t=1}^T \ln p(y_t | x_t, W_d)
		\label{Equ:ProblemFormulation:SupervisedLearning}
	\end{align}
where $T$ is the number of training examples. It is clear that the supervised learning problem requires us to label each $x_t$ with an output (label) $y_t$ in order to solve the above optimization problem \eqref{Equ:ProblemFormulation:SupervisedLearning}. 

In this paper, we consider the unsupervised learning of $p(y_t | x_t, W_d)$ from unpaired training \emph{sequences} $\{x_t, t=1,\ldots, T\}$ and $\{y_t, t=1,\ldots, T\}$. The input samples $\{x_t\}$ and the output samples $\{y_t\}$ are unpaired in that they are not necessarily generated from the true joint distribution $p(x_1,\ldots, x_T, y_1, \ldots, y_T)$ that we are trying to learn, and they are only required to be distributed according to the respective marginal distributions, i.e., $\{x_t\} \sim p(x_1,\ldots, x_T)$ and $\{y_t\} \sim p(y_1,\ldots, y_T)$. Therefore, $\{x_t\}$ and $\{y_t\}$ could be collected from two completely independent sources. In the rest of the paper, we assume that the probability distribution $p(y_1,\ldots, y_T)$ of the output samples has a sequence structure, i.e., there is temporal dependency over $y_1,\ldots, y_T$. Furthermore, we assume that $p(y_1,\ldots, y_T)$ is known a priori, which, as we pointed out earlier, could be estimated from a different data source that has the same distribution of $p(y_1,\ldots, y_T)$. 

More formally, our objective in this paper is to learn the posterior probability $p(y_t | x_t, W_d)$ (i.e., the predictor) from the input sequence $\{x_t\}$ by exploiting the distribution $p(y_1,\ldots, y_T)$ on the output sequence, where $p(y_1,\ldots, y_T)$ is learned from another totally unpaired sequence $\{y_1,\ldots, y_T\}$. Therefore, this is an unsupervised learning problem, which we will proceed to solve and analyze in the rest of the paper.

\section{Learning to Predict from Unpaired Samples}

We now develop a novel cost function for learning the predictor $p(y_t | x_t, W_d)$ in an unsupervised manner. The cost function is designed based on the following two key insights. First, given a predictor $p(y_t | x_t, W_d)$, we want the predicted output sequence $\hat{y}_1, \ldots, \hat{y}_T$ from the input sequence $x_1,\ldots,x_T$  to be consistent with the output distribution $p(y_1,\ldots, y_T)$, with the definition of consistency to be explained later. Second, we want the predicted output $\hat{y}_t$ to be based on the input $x_t$; that is the output $\hat{y}_t$ should be correlated with the input $x_t$ rather than completely independent of it. Therefore, our proposed cost function will have two terms. The first term measures how well the predicted output fit into the output distribution, and the second condition is a regularization term, which prevents the learning algorithm from overfitting into $p(y_1,\ldots, y_T)$ and obtaining trivial solutions that generate $\hat{y}_t$ completely independently of the input $x_t$.  Below, we formalize these ideas by developing these two terms in the cost function.

We first establish the first term in the novel unsupervised learning cost function. Note that, for each input sample $x_t$, the parametric conditional distribution $p(y_t | x_t, W_d)$ defines a probability of the corresponding output sample $y_t$. When the predictor $p(y_t | x_t, W_d)$ is applied to each sample in the input sequence $x_1,\ldots, x_T$, and generates the output according to this distribution, we will generate a random output sequence $\hat{y}_1,\ldots, \hat{y}_T$. Then, the log-likelihood $\ln p(\hat{y}_1,\ldots, \hat{y}_T)$ measures how well the generated sequence fit into the distribution $p(y_1,\ldots, y_T)$. Motivated by this observation, we define the following term to measure the expected fitness of the predicted output with the current predictor:
	\begin{align}
		\E\left[
			\ln p(y_1,\ldots, y_T) 
			\big|
			x_1,\ldots, x_T
		\right]
			&=
				\E\left[
					\sum_{t=1}^T \ln p(y_t | y_{t-1},\ldots, y_1) 
					\Big|
					x_t,\ldots, x_1
				\right]
				\nn\\
			&=
				\sum_{(y_{t},y_{t-1},\ldots,y_{1})}
				\prod_{t=1}^T p(y_t | x_t, W_d)
				\sum_{t=1}^T
				\ln p(y_t | y_{t-1}, \ldots, y_1) 
				\nn\\
			&=
				\sum_{t=1}^T
				\prod_{\tau=1}^{t-1}
				p(y_{\tau} | x_{\tau})
				\sum_{y_t}
				p(y_t | x_t) 
				\ln p(y_t | y_{t-1}, \ldots, y_1)
				\nn\\
			&=
				\sum_{t=1}^T
				\E\left[
					\sum_{y_t}
					p(y_t | x_t) 
					\ln p(y_t | y_{t-1}, \ldots, y_1)
					\Big|
					x_{t-1},\ldots, x_{1}
				\right]
		\label{Equ:UnsupLearn:Cost_FirstTerm}
	\end{align}
where the last expectation is evaluated with respect to $\prod_{\tau=1}^{t-1} p(y_{\tau} | x_{\tau}, W_d)$. The learning algorithm seeks to maximize the above objective function \eqref{Equ:UnsupLearn:Cost_FirstTerm} in order to make the predicted output sequence fit well into the prior distribution $p(y_1,\ldots, y_T)$. We will further show in the next section that the global optimal solution to \eqref{Equ:UnsupLearn:Cost_FirstTerm} is indeed the ground truth solution if the parametric model $p(y_t | x_t, W_d)$ includes the ground truth as one of its solution. 

However, we will further reveal in the next section that this objective function has many local optima that are badly behaved. These local optima lead to trivial solutions, which completely ignore the input data and produce outputs that fit into $p(y_1,\ldots, y_T)$. To address this issue, we introduce the second term in the cost function, which penalizes the solution that decouples the inputs and outputs. Specifically, we propose to use the following term
	\begin{align}
		\sum_{t=1}^T
		\E\Big[
			\ln p(x_t | y_t, W_g)
			| x_t
		\Big]
			&=
				\sum_{t=1}^T
				\sum_{y_t}
				p(y_t | x_t,W_d)
				\ln p(x_t | y_t, W_g)
		\label{Equ:UnsupLearn:Cost_SecondTerm}
	\end{align}
where $p(x_t | y_t, W_g)$ is a generative model parameterized by $W_g$ for characterizing the information flow from output to input. The expression \eqref{Equ:UnsupLearn:Cost_SecondTerm} has the following interpretation. For a given input sample $x_t$, we generate an output sample $y_t$ according to the distribution $p(y_t | x_t, W_d)$. Then for this particular sample $y_t$, the score $\ln p(x_t | y_t, W_g)$ measures how well the generative model $p(x_t | y_t, W_g)$ can predict the input $x_t$. During the learning process, we seek to maximize this term with respect to $W_g$ to maximize the generative model's ability to reconstruct the input from the output. That is, the learning process also learns the best generative model that can reconstruct the input from the output. 
	
Putting these two terms together, we have the following cost function for learning the predictor from unpaired data:
	\begin{align}
		\max_{W_d, W_g} \!
		\sum_{t=1}^T \!\!
		\left\{\!
			\E \! \left[\!
				\sum_{y_t}
				p(y_t | x_t) 
				\ln p(y_t | y_{t-1}, \ldots, y_1)
				\Big|
				x_{t-1},\ldots, x_{1} \!
			\right]
			\!\!+\!			
			\!
			\lambda \!
			\sum_{y_t}
			p(y_t | x_t,W_d)
			\ln p(x_t | y_t, W_g)
			\!\!
		\right\}
		\label{Equ:UnsupLearn:FinalCost}
	\end{align}
where $\lambda$ is a positive hyper-parameter that controls the relative ratio between the two terms. In the above optimization problem, we maximize the objective function with respect to both $W_d$ and $W_g$. As we discussed earlier, the maximization with respect to $W_g$ learns the best generative model to measure the ``correlation'' between the input and the predicted output from the discriminative model. Expression \eqref{Equ:UnsupLearn:Cost_SecondTerm} shows that this term also depends on $W_d$, which means that by maximizing \eqref{Equ:UnsupLearn:FinalCost} with respect to $W_d$, we are also maximizing the correlation between the input and the predicted output, thereby regularizing the learning of the discriminative model $p(y_t | x_t, W_d)$ to avoid trivial solutions.

The above learning problem \eqref{Equ:UnsupLearn:FinalCost} can be solved by using stochastic gradient, and the gradients can be computed by back propagation if the discriminative model $p(y_t | x_t, W_d)$ and the generative model $p(x_t | y_t, W_d)$ are (deep) neural networks.

\section{Experiments and Analysis}

In this section, we use a simplified prediction task on a synthetic dataset to study the effectiveness of the proposed approach. We will also analyze the behaviors of the proposed objective function in order to understand the nature and difficulties of the unsupervised learning problem for prediction along with its potential solutions.

\subsection{Experimental setup}

\begin{figure}[t]
\begin{center}
\includegraphics[width=0.4\textwidth]{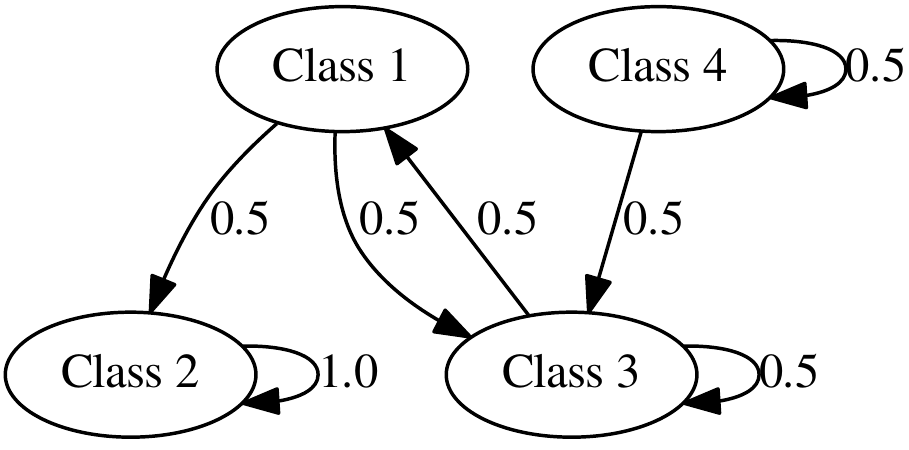}
\caption{\small{The transition probability of output observation.}}
\label{fig:markov_chain} 
\end{center}
\end{figure}

The synthetic data we use to evaluate the algorithm are generated in the following manner. We first generate the output sequence $y_1,\ldots, y_T$ according to the distribution $p(y_1,\ldots, y_T) = \prod_{t=1}^T p(y_t | y_{t-1})$, i.e., a Markov chain, which is described by Figure \ref{fig:markov_chain}. And we consider a four-class classification problem so that $y_t$ is a 4-dimensional one-hot vector. After the sequence $y_1,\ldots, y_T$ is generated, we randomly generate a permutation matrix $Q$ and fix it over time. For each $y_t$, we generate $x_t$ by multiplying $Q$ to the left of $x_t$, i.e., $x_t = Q y_t$. Therefore, the inputs $\{x_t\}$ are also a 4-dimensional one-hot vectors except that each of them is transformed from the output $y_t$ according to an unknown permutation. Our objective is to learn $p(y_t | x_t, W_d)$ from the input sequence $x_1, \ldots, x_T$ without the paired output sequence $y_1,\ldots, y_T$. Instead, we only have a sequence of unpaired samples $y_1,\ldots, y_T$ that is generated according to the same distribution $p(y_1,\ldots, y_T)$, from which we could estimate $p(y_1,\ldots, y_T)$. In our study below, we choose $p(y_t | x_t, W_d)$ and $p(x_t | y_t, W_g)$ to be the softmax functions:
	\begin{align}
		p(y_t | x_t, W_d)	=	\mathrm{softmax}(\gamma W_d x_t)
		\qquad
		p(x_t | y_t, W_d)	=	\mathrm{softmax}(\gamma W_g y_t)
		\label{Equ:Experiment:LinearModels}
	\end{align}
where $\gamma$ is a positive number that controls the sharpness of the softmax function. Even though we are using simple linear classifiers, as we proceed to reveal, the unsupervised learning cost is still highly non-convex and the problem remains difficult.

\subsection{The landscape of the proposed unsupervised cost function}

\begin{figure}
	\centerline{
	\subfigure[Unsupervised vs supervised costs]{\includegraphics[width=0.34\textwidth]{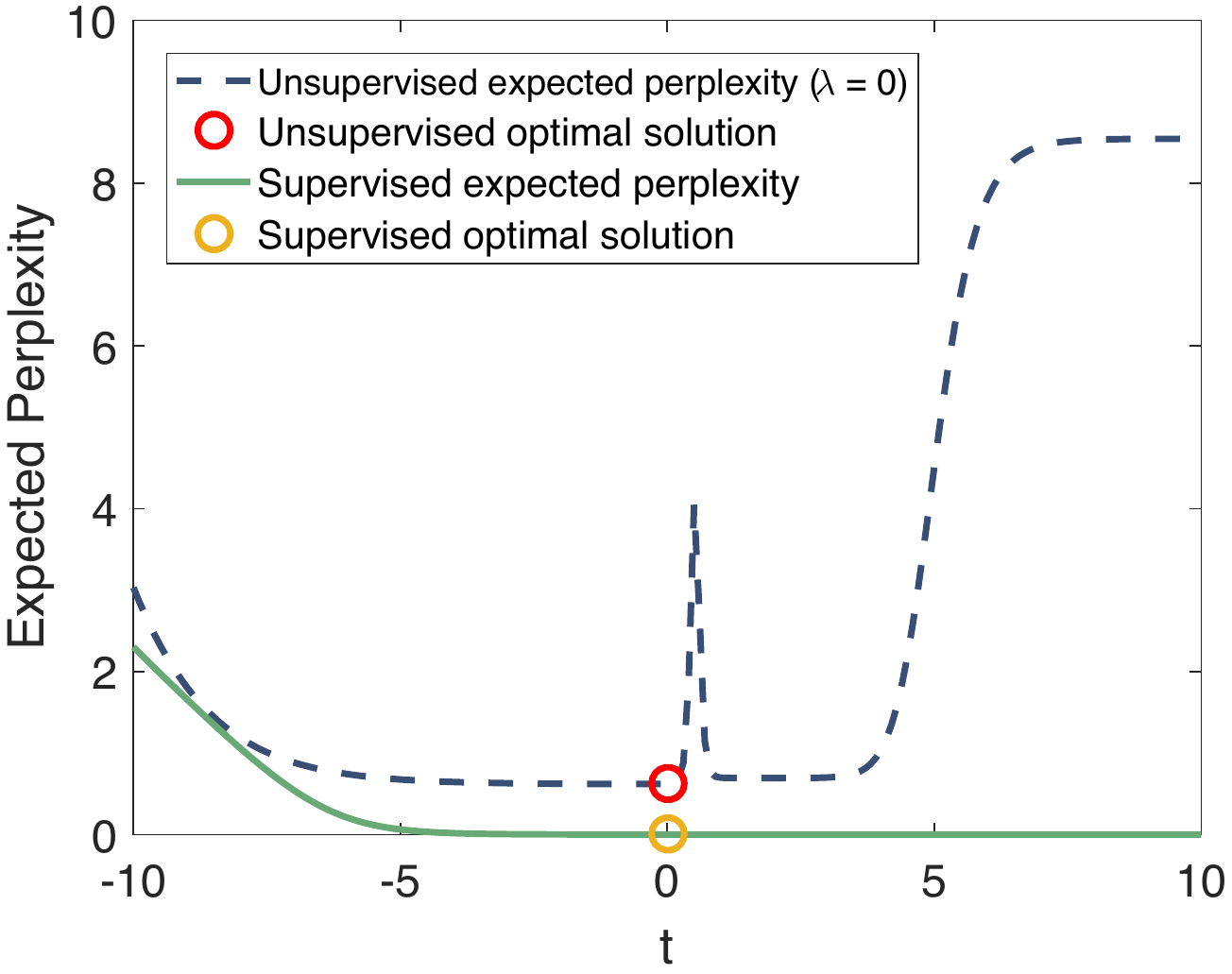}
	\label{fig:supervise_vs_unsupervise}
	}	
	\hfill
	\subfigure[The importance of regularization]{\includegraphics[width=0.35\textwidth]{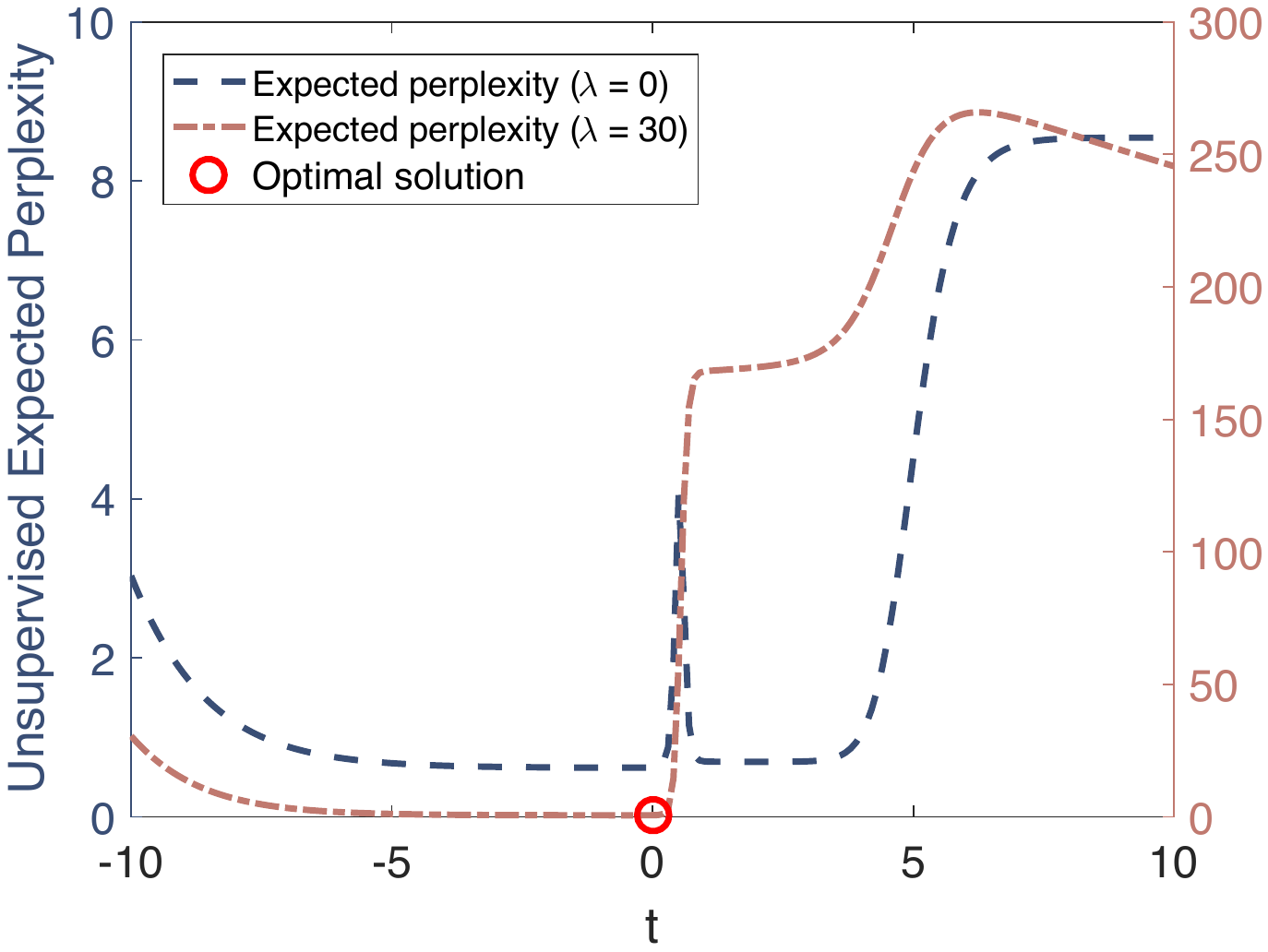}
	\label{fig:unsupervise_comparison}
	}		
	\hfill
	\subfigure[The local and global optima]{\includegraphics[width=0.32\textwidth]{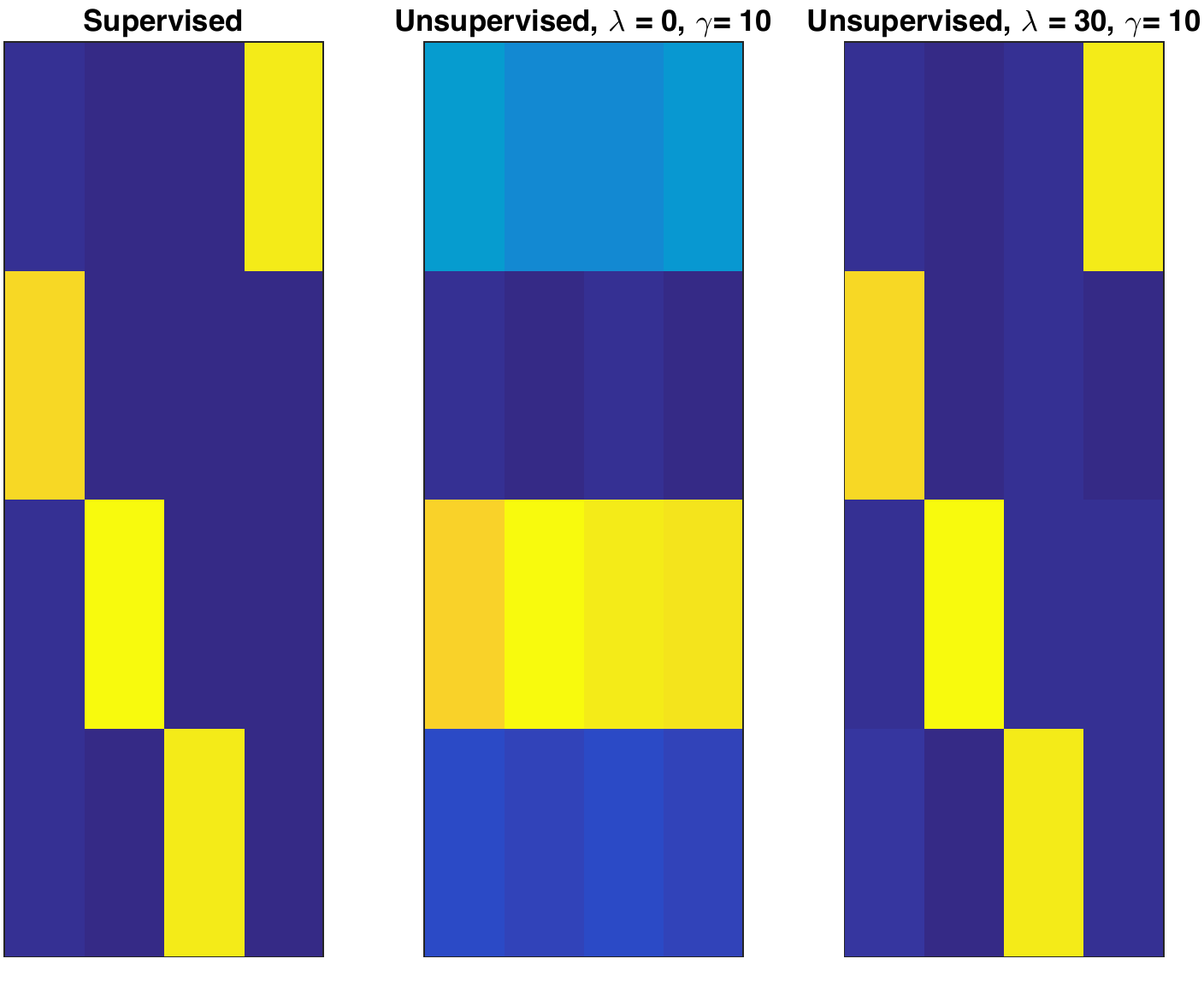}		\label{fig:unsupervise_rank_comparison}}
	}
	\caption{The landscape of supervised cost function, unsupervised cost functions (with different levels of regularizations), the local and global optimal solutions. Repeated experiments show similar results.}
	\label{Fig:UnsupervisedLandscapeAndSolutions}
\end{figure}

\begin{figure}
	\centerline{
	\subfigure[Unsupervised vs supervised costs (along random line 1)]{\includegraphics[width=0.42\textwidth]{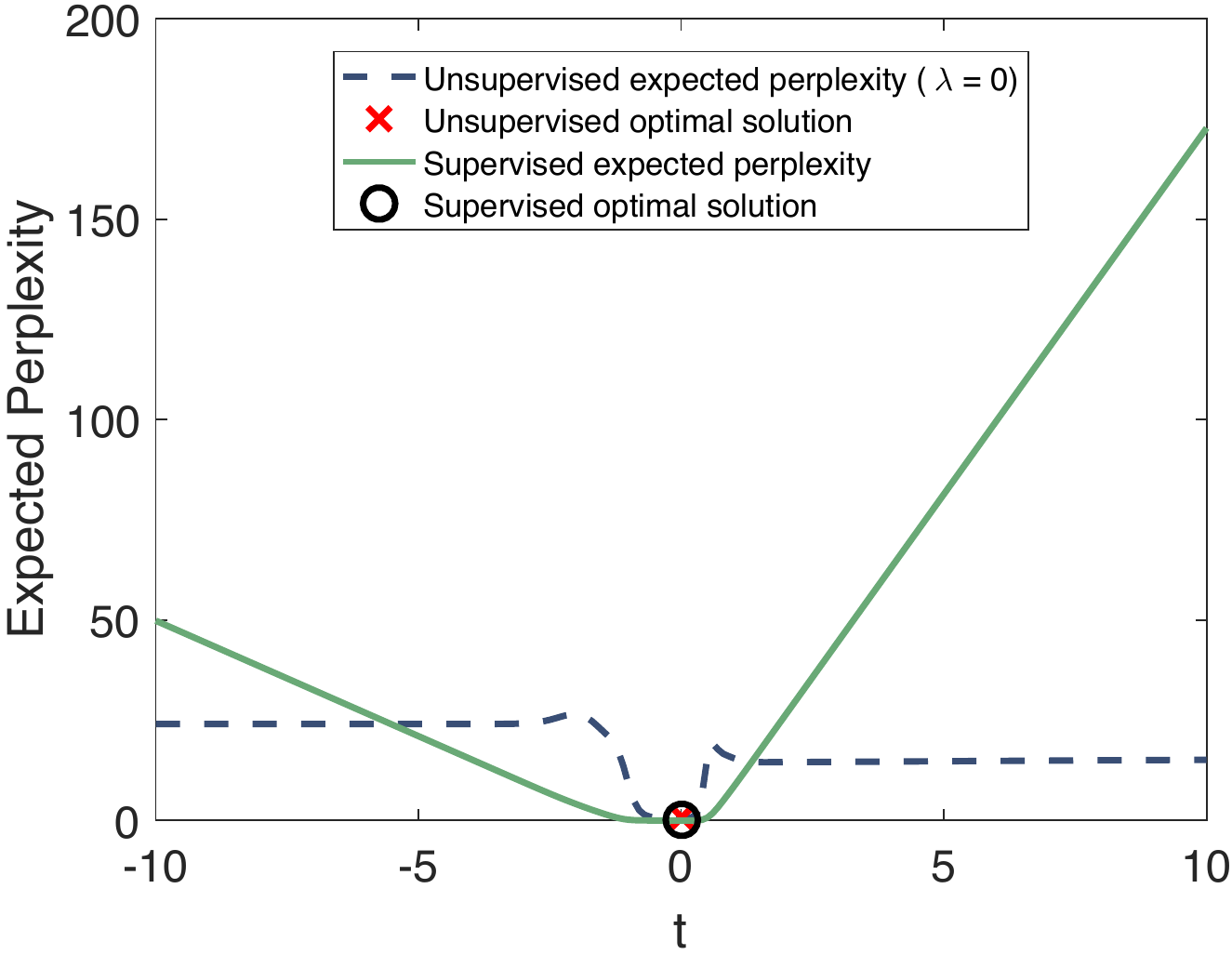}
	\label{fig:supervise_vs_unsupervise_random_1}
	}	
	\hfil
	\subfigure[The importance of regularization (along random line 1)]{\includegraphics[width=0.42\textwidth]{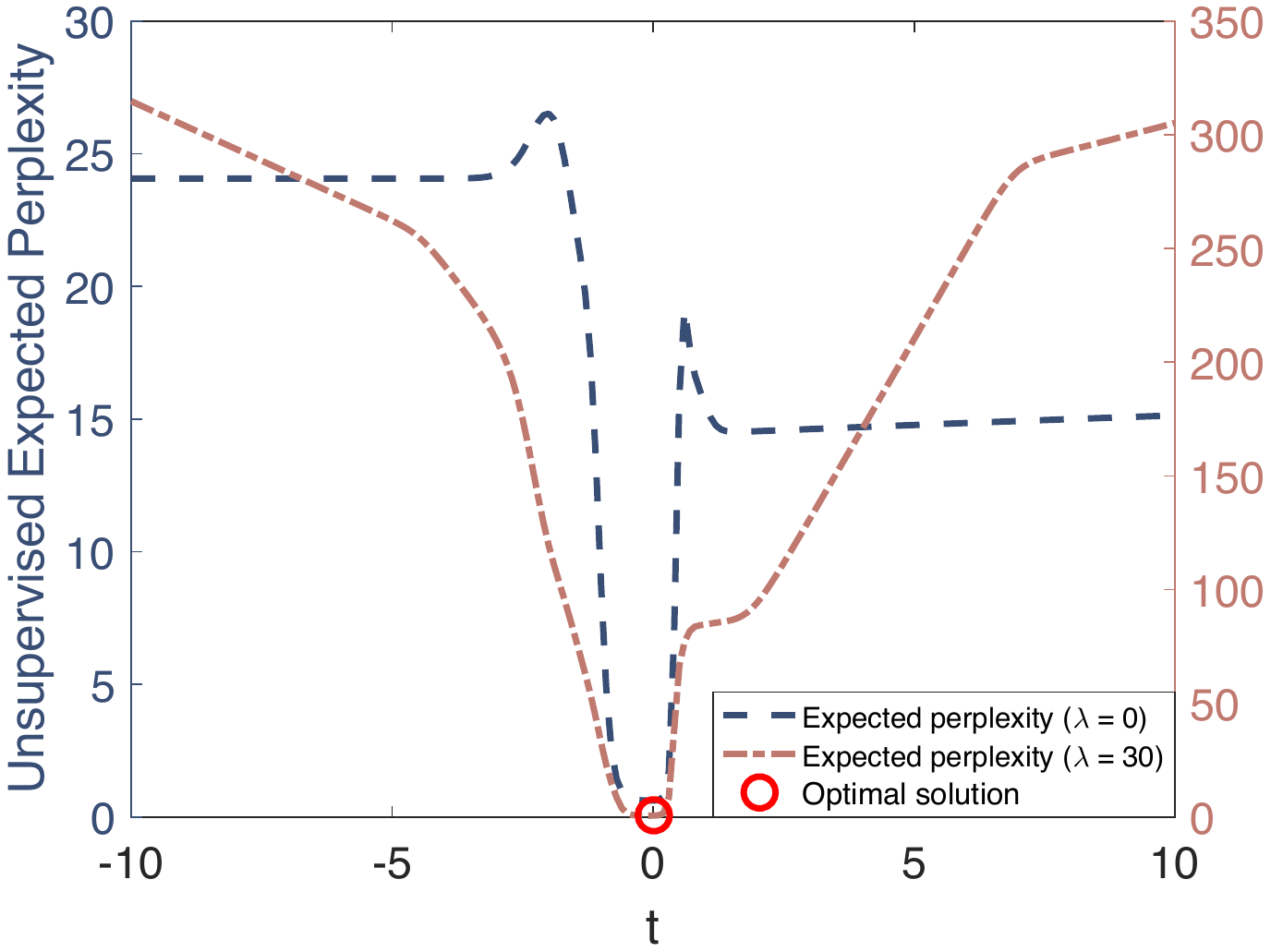}
	\label{fig:unsupervise_comparison_random_1}
	}
	}
	\centerline{
	\subfigure[Unsupervised vs supervised costs (along random line 2)]{\includegraphics[width=0.42\textwidth]{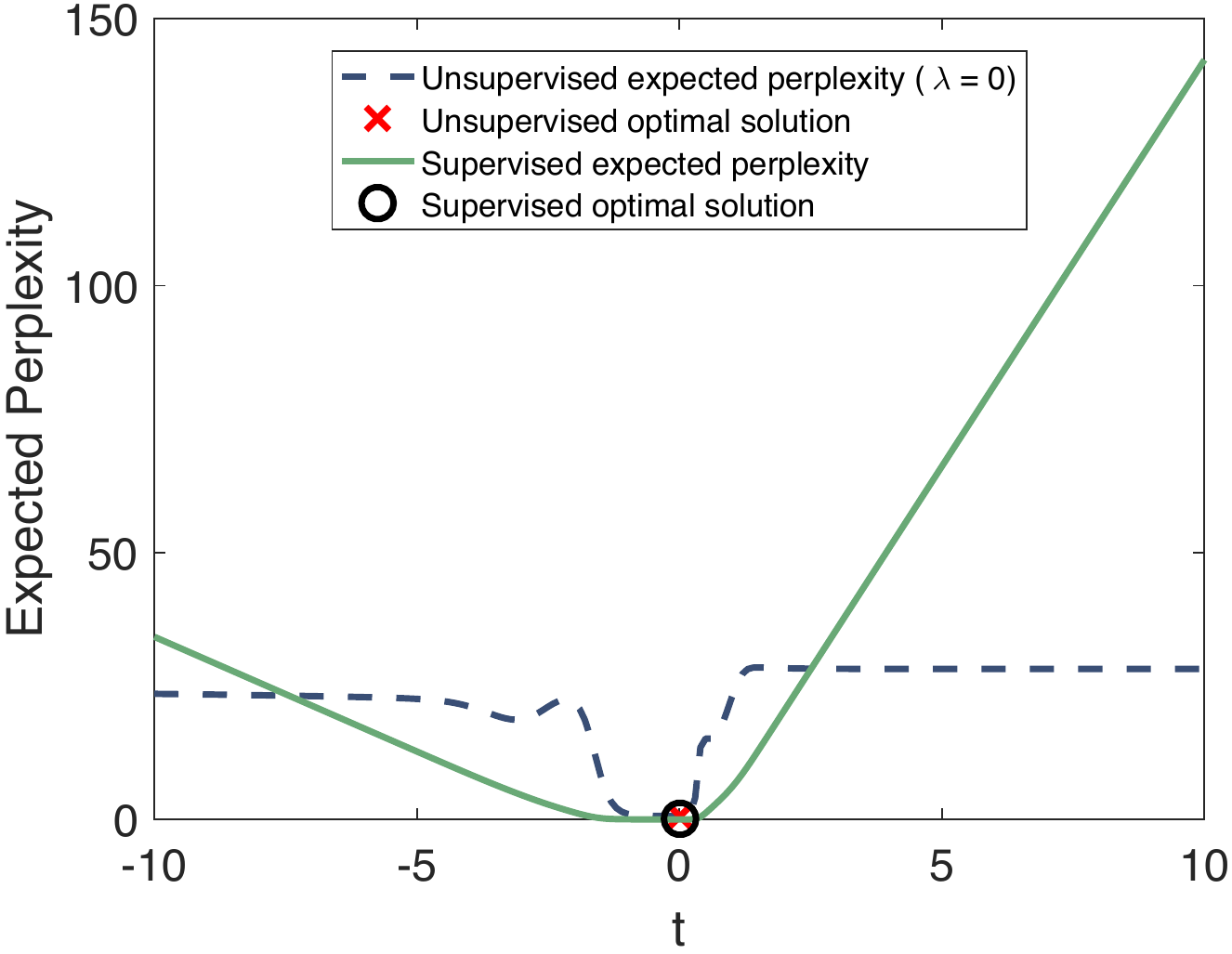}
	\label{fig:supervise_vs_unsupervise_random_5}
	}	
	\hfil
	\subfigure[The importance of regularization (along random line 2)]{\includegraphics[width=0.42\textwidth]{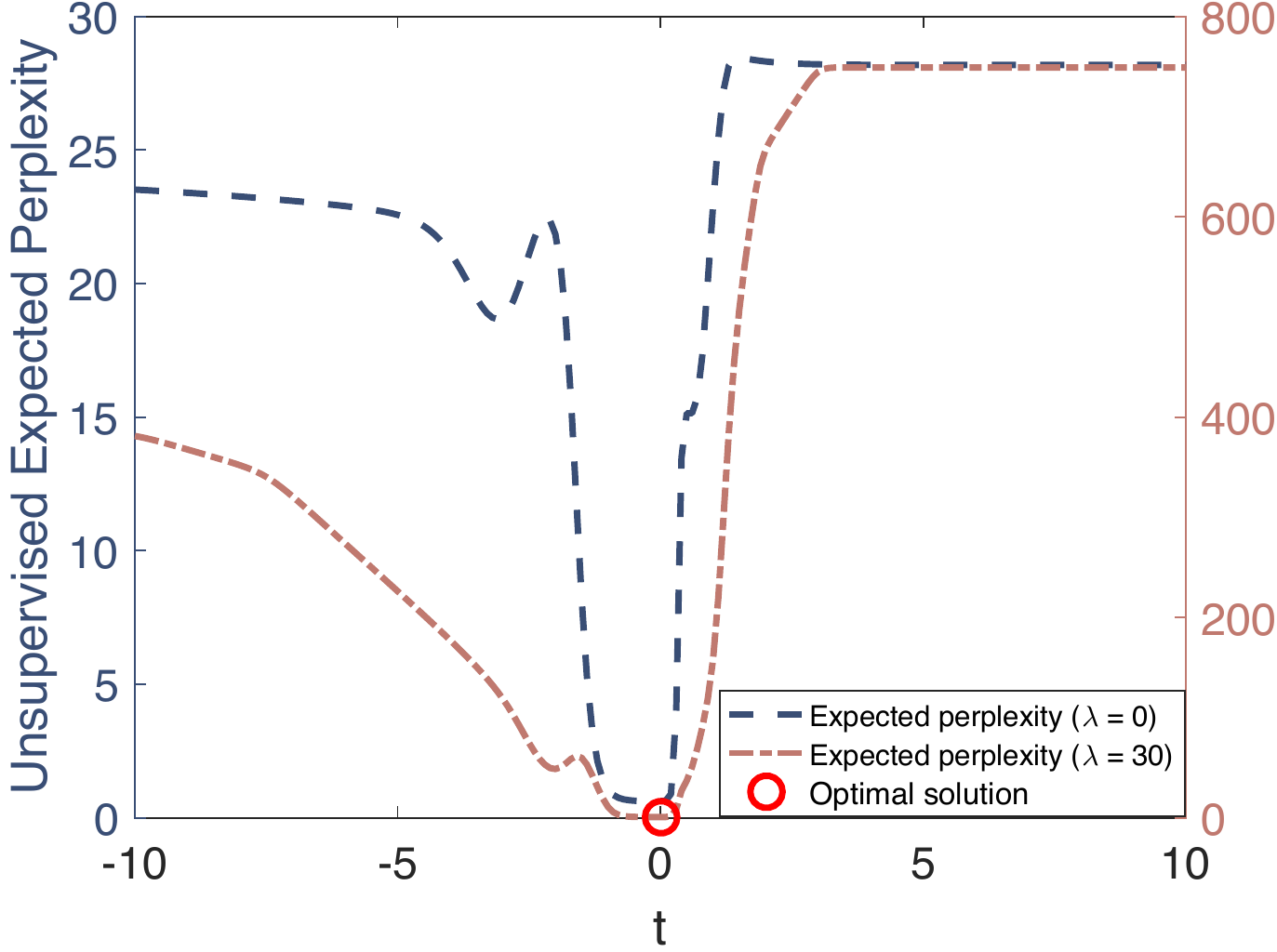}
	\label{fig:unsupervise_comparison_random_5}
	}
	}		
	\caption{The landscape of supervised cost function and unsupervised cost functions (with different levels of regularizations) along random lines that pass through the ground truth solution.}
	\label{Fig:UnsupervisedLandscapeAlongRandomLines}
\end{figure}

We first plot the landscape of the cost function \eqref{Equ:UnsupLearn:FinalCost} for $\lambda = 0$ case and compare it with the supervised cost (cross-entropy) in Figure \ref{fig:supervise_vs_unsupervise}.  Specifically, we plot the \emph{negative} of the objective function \eqref{Equ:UnsupLearn:FinalCost} along the line $t W_{d,0} + (1-t) W_{d,1}$, where $t$ is a real scalar, $W_{d,0}$ is the ground truth (obtained from the permutation matrix) and $W_{d,1}$ is the finally converged solution by optimizing \eqref{Equ:UnsupLearn:FinalCost} without regularization ($\lambda=0$). Obviously, the objective function is highly-nonconvex. On the other hand, the cost function for supervised learning is convex since the classifier is linear. An important observation we can make from Figure \ref{fig:supervise_vs_unsupervise} is that the global optimal solution to \eqref{Equ:UnsupLearn:Cost_FirstTerm} (i.e., the first term in \eqref{Equ:UnsupLearn:FinalCost}) coincides with the global optimal solution of the supervised learning problem. On the other hand, there is a local optimal solution, which the algorithm could easily get stuck in, as shown in the figure. We also note that the cost function of the local optimal solution seems to be very close to that of the global optimal solution. There are two important questions to ask: (i) how good is this local optimal solution in compare with the global optimal solution, and (ii) how does the regularization term (second term in \eqref{Equ:UnsupLearn:FinalCost}) help the algorithm escape from local optima. To answer the first question, we visualize the weight matrix $W_d$ in the middle part of Figure \ref{fig:unsupervise_rank_comparison}. We observe that the columns of the matrix are linearly dependent and the matrix is almost rank one by computing its singular values. With $W_d$ being rank-1 (e.g., $W_d \approx a b^T$), the probability $p(y_t | x_t, W_d) = \mathrm{softmax}(\gamma a b^T x_t) = \mathrm{softmax}(a)$, which is independent of $x_t$. Therefore, this local optimal solution is a trivial solution which totally ignores the inputs, although its cost is close to that of the global optimal solution. We repeated the experiments many times and all the local optimal solutions end up with rank-1. In Figures \ref{fig:supervise_vs_unsupervise_random_1} and \ref{fig:unsupervise_comparison_random_1}, we plot more landscapes of the supervised and unsupervised cost functions along other random lines that pass through the ground truth solution. From the figures, we note similar behaviors as in Figure \ref{Fig:UnsupervisedLandscapeAndSolutions}.

\subsection{The importance of regularization}

We now address the second question on the importance of regularization. In Figure \ref{fig:unsupervise_comparison}, we plot the landscapes of the unsupervised cost function \eqref{Equ:UnsupLearn:FinalCost} for $\lambda=0$ and $\lambda=30$. The landscapes show the values of the cost function along a random line that passes through the ground truth (global optimal solution). We observe that the regularization term creates a ``slope'' at the original position of the local optimal solution, which allows the algorithm to escape from the trivial solution. In Figures \ref{fig:unsupervise_comparison_random_1} and \ref{fig:unsupervise_comparison_random_5}, we plot more landscapes for the unsupervised cost with different levels of regularization and note similar behaviors, where the local optima are smoothed out by the regularization term. In the end, the obtained solution with $\lambda=30$ is shown in the right part of Figure \ref{fig:unsupervise_rank_comparison}. As a reference, we also put the global optimal solution to the supervised problem in the left part of Figure \ref{fig:unsupervise_rank_comparison}. We see that the solution obtained from unsupervised learning problem \eqref{Equ:UnsupLearn:FinalCost} with $\lambda=30$ is very close to the supervised solution.


\subsection{The impact of imperfect $p(y_1,\ldots, y_T)$}

\begin{figure}
\centering
\includegraphics[width=0.5\textwidth]{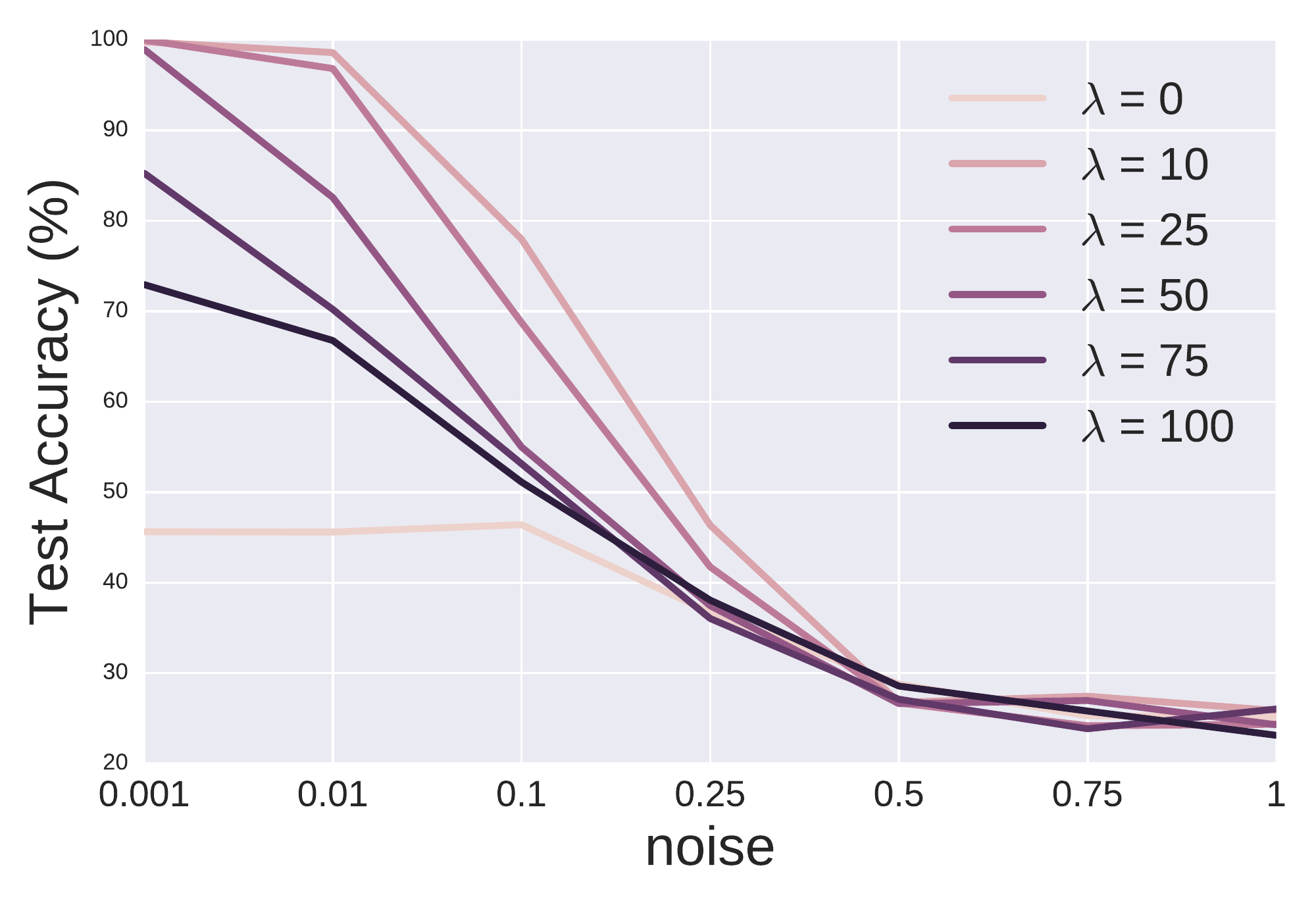}
\caption{Sensitivity of the performance with respect to the estimation accuracy of $p(y_1,\ldots, y_T)$.}
\label{Fig:6_transition_matrix_analysis}
\end{figure}

So far we have only considered the case where the probability $p(y_1,\ldots, y_T)$ is precisely known. In practice, this prior probability is estimated from a separate data sequence, which would always have estimation error. To examine the robustness of the algorithm with respect to the estimation error of $p(y_1,\ldots, y_T)$ (in this synthetic data case, $p(y_1,\ldots, y_T)$ is represented by the transition matrix $P$ of the Markov chain in Figure \ref{fig:markov_chain}), we add different levels of noise to the transition matrix be $P \leftarrow P+\mathcal{N}(0, \sigma_P^2)$ (and normalize the columns of $P$ so that they sum up to one) and evaluate the performance of the unsupervised learning algorithm. The test error for different variance of noise ($\sigma_P^2$) and different $\lambda$ are shown in Figure \ref{Fig:6_transition_matrix_analysis}. As the estimation error of $p(y_1,\ldots, y_T)$ increases, the performance of the unsupervised learning algorithm degrades. Furthermore, it is also noticeable that the regularization parameter $\lambda$ has to be set to a reasonable value to achieve the best performance. This is not surprising because if $\lambda$ is too small, the ``slope'' created by the regularization is not steep enough. On the other hand, if $\lambda$ is too large, the regularization term will overwhelm the first term (which contains the information regarding $p(y_t | x_t)$) in \eqref{Equ:UnsupLearn:FinalCost} so that the algorithm is not able to learn meaningful information.

%


\section{Conclusions}

In this paper we study the important problem of unsupervised learning for  prediction tasks, which is to learn to predict without using input-label paired data. We address this challenging problem by exploiting the sequence structure of the output samples to learn the predictor. That is, we proposed an objective function that aims to make the predicted outputs fit into the structure of the output while preserving the correlation between the input and the predicted output. On a synthetic structural prediction problem, we show that, even with simple linear classifiers, the objective function is already highly non-convex. On the other hand, this objective function converges to an optimal solution. We are currently investigating the behavior of more complicated and realistic models with real-world data.
 
Along this line of research, a recent work \cite{choromanska2014loss} shows that the local optima during supervised learning of the deep neural networks are well behaved. However, as we have demonstrated in our paper, this is not the case in the unsupervised learning problem, where the other local optimal solutions represent trivial solutions, although the values of the cost function are close to the global optimum. This leads to a further question on how to design even better objective functions to eliminate the trivial solutions from the set of local optima.

\small
\bibliography{NIPS2016_UnsupervisedLearning}
\bibliographystyle{plain}

%
%
%

\end{document}